\documentclass{article} %
\usepackage{iclr2026_conference,times}

\usepackage{amsmath,amsfonts,bm}

\def\eqref#1{equation~\ref{#1}}

\def\1{\bm{1}}

\DeclareMathAlphabet{\mathsfit}{\encodingdefault}{\sfdefault}{m}{sl}
\SetMathAlphabet{\mathsfit}{bold}{\encodingdefault}{\sfdefault}{bx}{n}

\usepackage[pagebackref=true,colorlinks,citecolor=brown]{hyperref}
\renewcommand{\paragraph}[1]{\vspace{.1em}\noindent\textbf{#1}}

\usepackage{url}

\usepackage{xspace}
\usepackage{mathtools,amsthm}
\usepackage{amsmath}
\usepackage{amssymb}
\usepackage{algorithm}
\usepackage[capitalize]{cleveref}
\usepackage{multirow}

\usepackage{algpseudocode}

\crefname{prop}{Proposition}{Propositions}

\newcommand{\xpm}[1]{{\tiny$\pm#1$}}

\usepackage{bm}
\usepackage{mathtools}

\usepackage{multirow}
\usepackage{booktabs}  
\usepackage{enumitem}

\usepackage[dvipsnames]{xcolor}
\newcommand{\plus}[1]{{\textcolor{Green}{$#1$}}}
\newcommand{\minus}[1]{{\textcolor{Red}{$#1$}}}
\title{Product of Experts for Visual Generation}

\author{%
  Yunzhi Zhang$^{1}$
  \quad Carson Murtuza-Lanier$^{1}$
  \quad Zizhang Li$^{1}$
  \quad Yilun Du$^{2}$
  \quad Jiajun Wu$^{1}$ 
  \\
$^1$Stanford University \quad $^2$Harvard University
}

\iclrfinalcopy %
\begin{document}

\maketitle

\begin{abstract}
Modern neural models capture rich priors and have complementary knowledge over shared data domains, e.g., images and videos. Integrating diverse knowledge from multiple sources—including visual generative models, visual language models, and sources with human-crafted knowledge such as graphics engines and physics simulators remains under-explored. We propose a probabilistic framework that combines information from these heterogeneous models, where expert models jointly shape a product distribution over outputs. To sample from this product distribution for controllable image/video synthesis tasks, we introduce an annealed  MCMC sampler in combination with SMC-style resampling to enable efficient inference-time model composition. Our framework empirically yields better controllability than monolithic methods and additionally provides flexible user interfaces for specifying visual generation goals. 
\footnote{Project page: {\footnotesize\url{https://product-of-experts.github.io/}}.}

\end{abstract}

\section{Introduction}
\label{sect:introduction}

Recent image and video generative models~\citep{saharia2022photorealistic,rombach2022high,ho2022imagen,videoworldsimulators2024} have achieved remarkable success in realistic appearance modeling, yet still have limitations in following complex text instructions and adhering to real-world constraints such as physical laws. 
The former would benefit from semantic priors in visual language models (VLMs)~\citep{radford2021learning,li2023blip,bai2023qwen,achiam2023gpt}, and the latter from rules embedded in physics simulators. 
However, training a single model to absorb all information sources (texts/visual corpora, simulation trajectories, etc.) can be prohibitively expensive. 

We address this challenge by integrating knowledge across models at inference time~\citep{du2024compositional}, leveraging visual generative priors, discriminative rewards from VLMs, and rule-based knowledge from physics simulators.
We focus on the task of controllable visual generation and aggregate opinions across a set of ``expert'' models by sampling from the product distribution defined by the models. Each expert, in our case, pretrained models such as VLMs or image and video generative models,  focuses on one or more constraints in generation, and the resulting product distribution naturally assigns high probabilities only to samples that satisfy all constraints simultaneously. The use of product distributions to combine the opinions of multiple experts has been extensively used in the past~\citep{bacharach1972scientific,genest1986combining, hinton1999products}, and in the visual domain in Markov Random Forests~\citep{wang2013markov, kolmogorov2002multi, glocker2008dense, boykov2006graph} and Conditional Random Fields~\citep{boykov2001interactive, kumar2003discriminative, he2004multiscale}.

Despite its conceptual appeal, sampling from product distributions is often intractable. Straightforward approaches like rejection sampling are inefficient due to vanishing acceptance rates in high dimensions. Recent works have gotten around these difficulties by either composing model distributions in a simpler Gaussian latent space~\citep{huang2022multimodal} or by composing through models through an annealed Langevin procedure~\citep{du2020compositional, du2023reduce, geffner2023compositional}, where the combination of annealing and gradient-based sampling enables the modes of the product probability distribution to be effectively identified. In this paper, we provide a more general framework that enables us to sample from the product of both autoregressive and diffusion-based generative models, as well as discriminative models such as VLMs. Our approach relies on a combination of Annealed Importance Sampling~\citep{neal2001annealed} and Sequential Monte Carlo~\citep{doucet2001introduction} to effectively find the modes of the product distribution across all experts.

Our overall framework enables us to integrate multiple input constraints without costly retraining.
We empirically validate its benefits in image generation following complex text instructions.
It also naturally yields a flexible user interface for defining complex generative goals by selecting and configuring experts, e.g., allowing users to specify the pose and motion trajectory of an object of interest, insert it into an existing image, and animate the full scene. 

Overall, our contributions are threefold: 
\begin{itemize}[leftmargin=2em, nosep]
    \item[{\bf i)}] We formulate controllable visual synthesis tasks in a unified Product-of-Experts (PoE) framework that enables principled knowledge integration of heterogeneous generative and discriminative models, as well as physics simulators. 
    \item[{\bf ii)}] We propose a practical, computationally efficient sampling framework based on Annealed Importance Sampling (AIS) and Sequential Monte Carlo (SMC).
    \item[{\bf iii)}] Our method enables flexible forms of user-instruction-following (texts, images, and low-level specifications such as object poses and trajectories) for image/video synthesis applications and achieves better user controllability and output fidelity compared to baseline methods. 
\end{itemize}

\section{Related Works}

\paragraph{Compositional Generative Modeling.} 
Prior works on compositional generative modeling typically combine multiple generative models to jointly generate data samples~\citep{du2020compositional, garipov2023compositional, du2023reduce, huang2022multimodal, mahajan2024compositional, du2024compositional, bradley2025mechanisms, thornton2025composition, gaudi2025coind} and has been previously applied in visual content generation~\citep{liu2022compositional,bar2023multidiffusion,zhang2023diffcollage,yang2024probabilistic,su2024compositional, li2022composing}, but such applications typically focus on multiple homogeneous generator experts, or, in the case of \cite{li2022composing}, a single generator expert with multiple discriminative experts. In contrast, our approach provides a general probabilistic framework through which many heterogeneous generative and discriminative experts can be jointly combined in generation. 

\paragraph{Reward Steering from Discriminative Models.} 
Our work is further related to recent work steering generative models with discriminative models and reward functions. Methods using gradient descent on a discriminative reward~\citep{grathwohl2019your, dhariwal2021diffusion, bansal2023universal,luo2025dual,he2023manifold, ye2024tfg,song2023loss,rout2025semantic} assume dense, differentiable gradients on the reward.
Another branch of work applies SMC for more exact reward steering~\citep{wu2023practical, zhao2024probabilistic,singhal2025general}. 
Both \cite{skreta2025fkc,he2025rne} and our work build on top of this formulation and consider composing both generative and discriminative experts, but with the following difference: when considering generative expert products (\cref{eqn:target-generative}), these works calculate \emph{path-wise} importance weights to ensure sampling correctness, risking weight degeneracy as path length grows~\citep{skreta2025fkc}, while ours uses \emph{per-timestep} MCMC kernels that leave $p_t$ (\cref{eqn:target-generative-annealed}) invariant without accumulating weights.

\paragraph{Video Generation with Physical Simulators.}
Our framework integrates knowledge from physical simulators to improve physical accuracy compared to end-to-end models, while alleviating the need for tedious 3D scene setups in traditional graphics rendering pipelines.
Recent works have explored such direction typically convert physical simulator outputs into a specific type of conditional signals such as optical flow~\citep{burgert2025go,montanaro2024motioncraft,yang2025towards} and point tracking~\citep{gu2025das}, or rely on a single pre-trained video generation model~\citep{liu2024physgen,chen2025physgen3d,tan2024physmotion}, while our framework allows for the integration of various signals available from physical simulation, providing more complete specifications of controls.

\section{Method}

We aim to generate complex scenes $x\in\mathbb{R}^d$ by combining the priors from both generative models (such as image/video models conditioned on texts or control signals converted from physics simulators) as well as discriminative models (such as VLMs), with examples detailed in \cref{sect:exp}. 
We formulate this composition of models in a probabilistic manner, where generative models are represented as probabilistic priors over data $\{p^{(i)}(x)\}_{i=1}^N$. Discriminative models are written as constraint functions, $\{r^{(j)}: \mathbb{R}^d \rightarrow \mathbb{R}\}_{j=1}^M$, each assigning a scalar reward that represents how much a sample $x$ matches the constraint encoded in the model. Each constraint function is converted into an unnormalized probability distribution through the Boltzmann distribution, $q^{(j)}(x) := (\exp{{r^{(j)}}})(x)$.

To combine these two classes of models, we aim to sample from the product distribution:
\begin{equation}
    x \sim p(x) :\propto \prod_{i=1}^N p^{(i)}(x)\prod_{j=1}^M q^{(j)}(x), 
    \label{eqn:target-distribution}
\end{equation}
where each generative expert $p^{(i)}(x)$ and discriminative expert $q^{(j)}(x)$ are defined over parts of the scene $x$. The product distribution $p(x)$ has high probability for a sample $x$ when $x$ has high probability under both $p^{(i)}(x)$ and  $q^{(j)}(x)$~\citep{hinton2002training,du2020compositional, du2024compositional}, allowing us to compose together the priors in both generative and discriminative models.

However, sampling from \cref{eqn:target-distribution} is challenging, as we need to search for samples that satisfies constraints across all experts. Below, we will discuss how we can efficiently and effectively sample from the \cref{eqn:target-distribution} for high fidelity visual generation. We first discuss how we can sample from multiple generative experts in \cref{sect:generative_expert} and then discuss how to sample jointly from multiple discriminative and generative experts in \cref{sect:generative_discriminative}, followed by a practical implementation in \cref{sect:method-instantiation}.

\subsection{Sampling from Generative Experts}
\label{sect:generative_expert}

We first discuss how we can effectively sample from the product of a set of generative experts,
\begin{equation}
    x \sim p(x) :\propto \prod_{i=1}^N p^{(i)}(x). 
    \label{eqn:target-generative}
\end{equation}
To sample from the above distribution, we can use Markov chain Monte Carlo (MCMC)~\cite{robert1999monte}, where we iteratively refine
samples based on their likelihood under the product distribution. 
In discrete settings, a variant of Gibbs sampling
may be used, while in continuous domains, Langevin sampling based on the gradient of log density may be used~\citep{welling2011bayesian}.

A central issue, however, is that MCMC is a local refinement procedure and can take an exponentially long time for the sampling procedure to mix and find a high likelihood scene $x$ from the product distribution~\citep{robert1999monte}. To effectively sample from product of generative experts, we propose to use Annealed Importance Sampling (AIS)~\citep{neal2001annealed} and construct of a sequence of $T$ distributions, $\{p_t(x)\}_{t=T}^1$, where $p_T(x)$ is a smooth, easy-to-sample distribution and $p_1(x)$ is the desired product distribution defined in \cref{eqn:target-generative}. To draw a sample from $p_1(x)$, we first initialize a sample from $p_T(x)$ and iteratively run MCMC on the sample to sample from each intermediate distribution before finally reaching $p_1(x)$. Since intermediate distributions are easier to sample from, this enables us to more effectively find a high likelihood scene $x$ from the product distribution.

We construct the intermediate probability distributions of the following form: 
\begin{equation}
    p_t(x) :\propto \prod_{i=1}^N p^{(i)}_t(x), 
    \label{eqn:target-generative-annealed}
\end{equation}
where $p^{(i)}_t(x)$ is an expert-specific distribution interpolating between an initial distribution (e.g., Gaussian or uniform) and the final expert distribution $p^{(i)}(x)$. We list concrete choices of $p^{(i)}_t(x)$ for two common classes of generative models in \cref{sect:method-instantiation}. For autoregressive models~\citep{oord2016pixel,larochelle2011neural}, it is the marginal distribution for a prefixed data region determined by $t$; for diffusion/flow models~\citep{liu2022flow,lipman2022flow,ho2020denoising,sohl2015deep}, it is the time marginal distribution with estimated score functions. 

\paragraph{Composing Conditional Generative Experts.} To improve the efficacy of sampling from the product distribution of generative experts, we can modify each generative expert to be conditionally dependent on the outputs of other experts. Specifically, we can represent the product distribution as
\begin{equation}
    x \sim p(x) :\propto \prod_{i=1}^N p^{(i)}(x_i|x_{\text{pa}(i)}), 
    \label{eqn:target-generative-conditional}
\end{equation}
where each $x_i$ refers to the part of a scene $x$ represented by the generative expert $p^{(i)}$, and $x_{\text{pa}(i)}$ refers to the other parts of a scene specified by other experts the expert is conditioned on. 

Making each generative expert conditionally dependent on the values of other experts reduces the multi-modality in the distribution $p^{(i)}(x)$, enabling more efficient MCMC sampling on the product distribution. Its benefit is empirically shown in \cref{fig:exp-image-edits-comparisons} and \cref{tab:editing}.

\subsection{Parallel Sampling with Discriminative Experts}
\label{sect:generative_discriminative}

\begin{algorithm}[t]
\caption{Product Distribution Sampling}
\begin{algorithmic}[1]
\label{alg:implementation}
\State \textbf{Input:} Annealing length $T$, initial distribution $p_T(\cdot)$, particle count $L$, MCMC step count $K$, generative experts $\{p_t^{(i)}(\cdot)\}_{i=1, t=1}^{N, T}$ with kernels $\mathcal{K}_{t\leftarrow t+1}(\cdot\mid\cdot)$ for MCMC initialization and $\mathcal{K}_{t}(\cdot\mid\cdot)$ for MCMC steps, discriminative experts with log probabilities $\{r^{(j)}(\cdot)\}_{j=1}^{M}$.
\State \textbf{Initialize:} Sample $x^{(l)} \sim p_T(\cdot)$, $\forall l = 1, \cdots, L$.
\For{$t = T-1$ to $1$}\Comment{Transition to the next annealed distribution}
    \For{$l=1$ to $L$}  \Comment{Parallel sampling}
        \State 
        $x^{(l)} \gets \mathcal{K}_{t\leftarrow t+1}(\cdot \mid x^{(l)})$ 
        \Comment{MCMC initialization}
        \For{$k=1$ to $K$}
            \State $x^{(l)} \gets \mathcal{K}_{t}(\cdot \mid x^{(l)})$
            \Comment{MCMC step}
        \EndFor
    \EndFor
    \State Resample samples with log weights $\sum_{j=1}^M r_t^{(j)}(x^{(l)})$ where $r_t^{(j)}(x^{(l)})\approx r^{(j)}(\hat{x}^{(l)})$.
\EndFor
\State $l^* \gets \max_l \sum_{j=1}^M r^{(j)}(x^{(l)})$
\State \textbf{Output:} $x^{(l^*)}$
\end{algorithmic}
\end{algorithm}

Next, we discuss how we can modify the annealed sampling procedure in Section~\ref{sect:generative_expert} to sample from the full product distribution from \cref{eqn:target-distribution}, including discriminative experts.
Similar to the previous section, we can define a sequence of intermediate sampling distributions 
\begin{equation}
    p_t(x) :\propto \prod_{i=1}^N p^{(i)}_t(x) \prod_{j=1}^M q^{(j)}_t(x). 
    \label{eqn:target-discriminative-annealed}
\end{equation}
To implement AIS, one simple approach is importance sampling as follows. We first obtain $L$ samples from the product of the intermediate generative expert distributions defined in \cref{eqn:target-generative-annealed} according to \cref{sect:generative_expert}, and then draw a weighted random sample from these $L$ samples, each with importance weight $\prod_{j=1}^M q^{(j)}_t(x)$ corresponding to the likelihood under product of discriminative experts.

An issue with this simple procedure is that each sample drawn from the MCMC may be correlated with each other, since MCMC is slow at mixing and may not cover the entire generative product distribution.  
To improve coverage, we use a parallel SMC sampler~\citep{doucet2001sequential}, where we maintain $L$ particles over the course of annealing. At each intermediate distribution, we run MCMC on each particle to draw samples from the product of the generative distributions, and then weigh and resample particles based on their likelihood under the product of the discriminative experts. 

Our overall algorithm requires only black-box access to each discriminator, requiring knowing only the likelihood the discriminator assigns to a sample $x$ and not additional information such as log-likelihood gradients. If we do have more information about the discriminative expert's form, we can directly treat it as generative as in \cref{sect:generative_expert} and use approaches such as gradient-based MCMC.

\begin{figure}
    \centering
    \includegraphics[width=\linewidth]{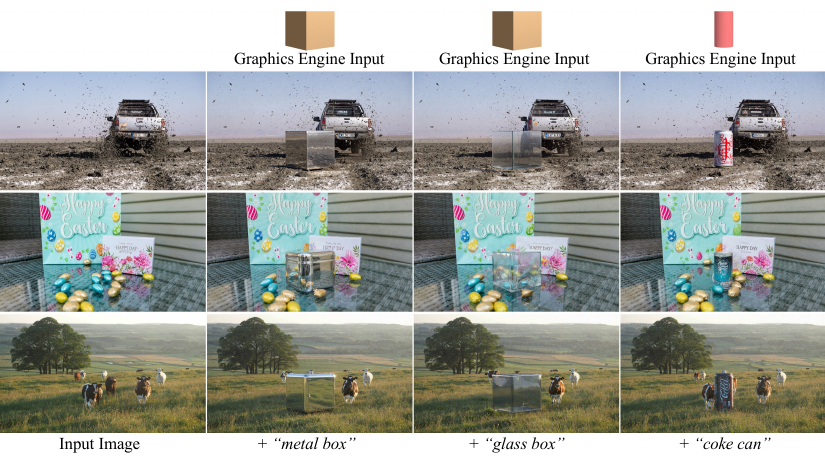}
    \vspace{-2.5em}
    \caption{\textbf{Application on Image Object Insertion} where the goal is to insert assets posed in a graphics engine (top row) and described with text prompts (bottom row) into images (first column).}
    \label{fig:exp-image-edits}
    \vspace{-2.2em}
\end{figure}

\subsection{Framework Instantiation}
\label{sect:method-instantiation}

We now introduce a specific implementation of the framework as summarized in \cref{alg:implementation}, which describes the annealed distribution and expert instantiations primarily used in the paper.

\paragraph{Generative Experts.}
The framework requires specifying the annealed generative distributions factors $\{p_t^{(i)}(x)\}_{t=T}^1$ (\cref{eqn:target-generative-annealed}), a kernel  $\mathcal{K}_{t\leftarrow t+1}(\cdot\mid \cdot)$ to propagate samples from the previous annealed distribution $p_{t+1}(x)$ to the next distribution $p_t(x)$ (line 5, \cref{alg:implementation}), and a MCMC kernel $\mathcal{K}_t(\cdot \mid \cdot)$ that leaves $p_t$ invariant (line 7). While $\mathcal{K}_{t\leftarrow t+1}$ does not draw a sample from $p_t(x)$ exactly, it serves as an effective initialization for MCMC sampling.

When generative expert $p^{(i)}(x)$ is a flow model, a natural choice for $\{p_t^{(i)}(x)\}_{t=T}^1$ is a discretization of the probability path generated by its velocity prediction $v^{(i)}_t(x)$. We set $\mathcal{K}_{t\leftarrow t+1}$ as one Euler step in the flow ODE integration, and $\mathcal{K}_t$ is the Langevin dynamics transition, both under composed score $\sum_i \nabla_x \log p_t^{(i)}(x)$ (\cref{app:framework-details-flow}). When $p^{(i)}$ is an autoregressive model, we use its prefix marginal for annealing, namely $p_t^{(i)}(x) = p^{(i)}(x_{1:T+1-t})$ such that $x_{1:T} = x$, where $x_{1:T+1-t}$ is a data slice. We set $\mathcal{K}_{t\leftarrow t+1}$ such that it appends the next data slice sampled from the model-predicted conditional distribution, and $\mathcal{K}_t$ can be implemented as Gibbs sampling (\cref{app:framework-details-autoregressive}).

\paragraph{Conditional Sampling.} We implement conditional generative experts $p_t^{(i)}(x_i\mid x_{\text{pa}(i)})$ for flow models by modifying the original generative experts $p_t^{(i)}(x_i)$. In particular,
we modify the generative flow $v^{(i)}_t(x_i)$ to take into account the predicted flow at parent regions
\begin{equation}
    v_t^{(i)}(x_i\mid x_{i^\prime}) \approx v^{(i)}_t(x_i)- w \sum_{i^\prime\in \text{pa}(i)} \nabla_{x_i} \| v^{(i)}_t(x_i) - \text{stopgrad}(v^{(i^\prime)}_t(x_{i^\prime}))\|_2^2,
    \label{eqn:conditional-update}
\end{equation}
where $w$ is the learning rate. Justifications of this approximation are deferred to \cref{app:conditional-sampling}.

\paragraph{Discriminative Experts.} 
We use pre-trained VLMs as reward functions $r^{(i)}(x)$. When generative experts are flow models, the noisy samples $x$ are out of distribution for discriminative experts. We define the intermediate counterparts in \cref{eqn:target-discriminative-annealed} as $r_t^{(j)}(x) = r^{(j)}(\hat{x})$ where $\hat{x}$ is the predicted endpoint of $x$~\citep{chung2022diffusion,efron2011tweedie} whose computation is specified in \cref{eqn:flow-endpoint}.

\begin{figure}
    \centering
    \includegraphics[width=\linewidth]{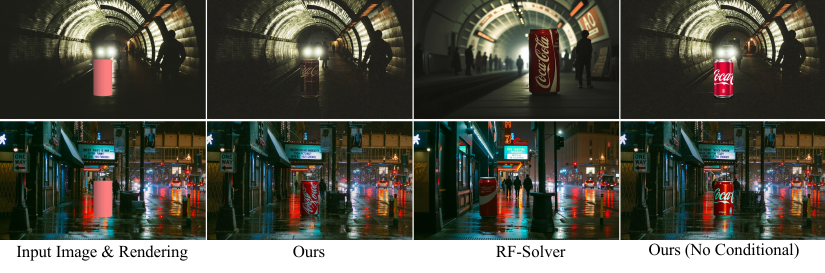}
    \vspace{-2.8em}
    \caption{\textbf{Image Object Insertion Comparisons.} Our method better adheres to input geometric conditions while faithfully preserving background details. The last column shows that conditional sampling improves visual harmonization and fidelity. 
    }
    \label{fig:exp-image-edits-comparisons}
    \vspace{-1.2em}
\end{figure}
\begin{table}[t]
\centering
\scriptsize
\setlength{\tabcolsep}{3pt}
\begin{tabular}{lccccccc}
\toprule
\multirow{2}{*}{Methods} & \multicolumn{3}{c}{Controllability} & \multicolumn{2}{c}{Image Quality} &
 \multicolumn{2}{c}{Text Alignment} \\
\cmidrule(lr){2-4} \cmidrule(lr){5-6}\cmidrule(lr){7-8}
 & MSE (bg) ($\downarrow$) & LPIPS (fg) ($\downarrow$) & GPT-4o ($\uparrow$) & Aesthetic ($\uparrow$) & GPT-4o ($\uparrow$) & ImageReward ($\uparrow$) & GPT-4o ($\uparrow$) \\
\midrule
\multicolumn{8}{l}{\textit{Graphics Engine Rendering Input}} \\
\midrule
RF-Solver
 & $1.619$\xpm{0.605}
 & $\underline{0.178}$\xpm{0.061}
 & $0.518$\xpm{0.168}
 & $\mathbf{0.618}$\xpm{0.070}
 & $0.662$\xpm{0.103}
 & $0.948$\xpm{0.777}
 & $0.438$\xpm{0.179} \\
Ours No Cond
 & $\underline{1.511}$\xpm{0.661}
 & $\mathbf{0.065}$\xpm{0.018}
 & $\underline{0.727}$\xpm{0.141}
 & $\underline{0.599}$\xpm{0.059}
 & $\underline{0.718}$\xpm{0.120}
 & $\underline{1.142}$\xpm{0.668}
 & $\underline{0.675}$\xpm{0.180} \\
Ours
 & $\mathbf{1.429}$\xpm{0.602}
 & $\mathbf{0.065}$\xpm{0.019}
 & $\mathbf{0.827}$\xpm{0.073}
 & $0.596$\xpm{0.061}
 & $\mathbf{0.783}$\xpm{0.082}
 & $\mathbf{1.175}$\xpm{0.666}
 & $\mathbf{0.817}$\xpm{0.129} \\
\midrule
\multicolumn{8}{l}{\textit{Character Rendering Input} (Magic Insert~\citep{ruiz2024magic} Dataset)} \\
\midrule
Magic Insert
 & $0.769$\xpm{0.535}
 & $0.106$\xpm{0.042}
 & $0.743$\xpm{0.142}
 & $0.721$\xpm{0.076}
 & $0.758$\xpm{0.130}
 & $1.169$\xpm{0.701}
 & $0.649$\xpm{0.141} \\
Add-it
 & $0.710$\xpm{0.473}
 & $0.128$\xpm{0.048}
 & \underline{$0.763$}\xpm{0.126}
 & $0.715$\xpm{0.066}
 & \underline{$0.778$}\xpm{0.097}
 & $1.078$\xpm{0.708}
 & \underline{$0.684$}\xpm{0.155} \\
FLUX-Fill
 & $\mathbf{0.058}$\xpm{0.044}
 & $0.103$\xpm{0.042}
 & $0.655$\xpm{0.147}
 & $0.691$\xpm{0.070}
 & $0.746$\xpm{0.092}
 & $0.828$\xpm{0.777}
 & $0.590$\xpm{0.152} \\
SDEdit
 & $0.968$\xpm{0.965}
 & $\mathbf{0.026}$\xpm{0.020}
 & $0.744$\xpm{0.131}
 & \underline{$0.724$}\xpm{0.074}
 & $\mathbf{0.871}$\xpm{0.061}
 & \underline{$1.640$}\xpm{0.376}
 & $0.658$\xpm{0.144} \\
Ours
 & \underline{$0.365$}\xpm{0.247}
 & \underline{$0.064$}\xpm{0.033}
 & $\mathbf{0.818}$\xpm{0.097}
 & $\mathbf{0.760}$\xpm{0.060}
 & $0.769$\xpm{0.118}
 & $\mathbf{1.711}$\xpm{0.308}
 & $\mathbf{0.763}$\xpm{0.107} \\
\bottomrule
\end{tabular}
\vspace{-1.5em}
\caption{\textbf{Image Object Insertion Evaluation.} 
}
\vspace{-3.2em}
\label{tab:editing}
\end{table}

\begin{figure}[t]
    \centering
    \includegraphics[width=\linewidth]{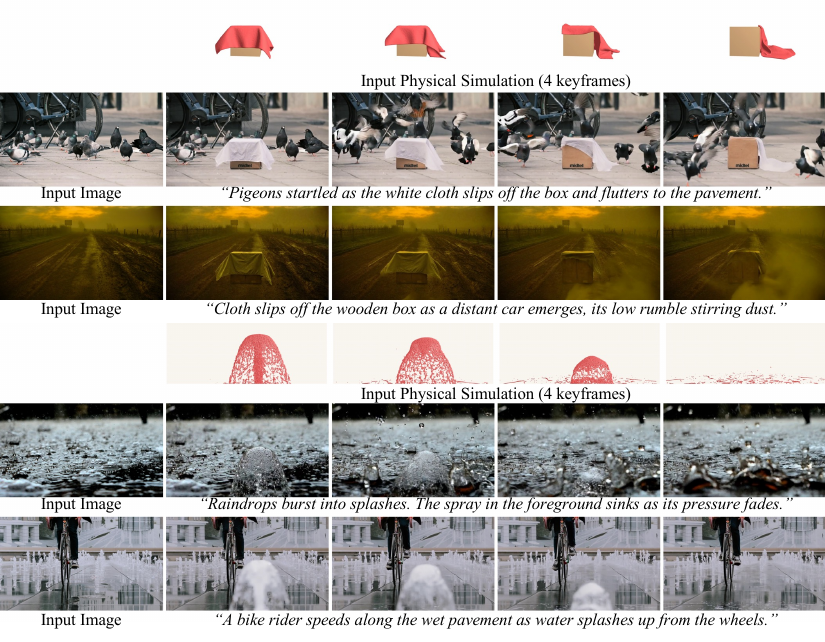}
    \vspace{-2.8em}
    \caption{\textbf{Application on Physical-Simulation-Instructed Video Generation.} Given an input image and a physical simulator rendering describing precise object motions, our method generates videos aligned with input motions while synthesizing natural content for non-foreground regions.}
    \vspace{-1em}
    \label{fig:exp-video}
\end{figure}

\section{Experiments}
\label{sect:exp}

We apply our framework to image/video synthesis tasks. Expert configurations can adapt to the desired, task-dependent input granularity, including high-level text instructions (\cref{sect:exp-editing,sect:exp-video,sect:exp-image}) and precise, low-level pose controls (\cref{sect:exp-editing,sect:exp-video}), and leverage expert knowledge ranging across natural image and video priors (\cref{sect:exp-editing,sect:exp-video,sect:exp-image}), precise physics rules from physical simulators (\cref{sect:exp-video}), and semantic visual understandings from VLMs (\cref{sect:exp-image}).

\subsection{Graphics-Engine-Instructed Image Editing}
\label{sect:exp-editing}

\vspace{-5pt}
\paragraph{Task.}
This task provides the following image object insertion interface for applications where users want to insert an object precisely into certain parts of the image. Inputs consist of an image to be edited, an input 3D asset posed in a graphics engine, and a text description describing higher-level information such as object materials (``metal'') and semantics (``coke can''). 
We deploy two generative experts for this task: a depth-to-image model, FLUX.1 Depth [dev], and an inpainting model, FLUX.1 Fill [dev]~\citep{flux2024}. The former ensures inserted objects follow input pose specifications, and the latter provides a natural image prior to produce realistic outputs.

\paragraph{Evaluation.}
We compare with an editing method, RF-Solver~\citep{wang2024rfsolver}, which inverts the input image to a Gaussian noise and then generates the output starting from that noise, conditioned on a text prompt that additionally describes where to insert the object, e.g., ``a metal box standing on the
ground''. The evaluation dataset consists of 10 natural images paired with 3 object assets, resulting in 30 scenes in total, with examples in \cref{fig:exp-image-edits}. The metrics cover three major aspects: controllability, image quality, and text alignment.
Specifically, we evaluate the MSE distance between generation outputs and input images on background pixels to evaluate background preservation, and LPIPS~\citep{zhang2018perceptual} distance between outputs and graphics engine's RGB renderings on foreground pixels to measure the fidelity to input objects. We report aesthetic score using the LAION-Aesthetic predictor~\citep{schuhmann2022laion} and measure text alignment with ImageReward~\citep{xu2023imagereward}, and query GPT-4o for automatic evaluation (details in \cref{app:exp-details-evaluation}). 

Results are in \cref{tab:editing} (top half). 
Baseline outputs do not conform to input object pose due to the lack of precision of text prompts, and may fail to preserve input background as observed in \cref{fig:exp-image-edits-comparisons}. 

\paragraph{Ablation on Conditional Sampling.}
\cref{sect:generative_expert} introduced the conditionally sampling strategy, which is crucial for efficient sampling from the product distribution. Removing the updates in \cref{eqn:conditional-update} (``No Cond'' in \cref{tab:editing}) decreases performance and results in less harmonized visual outputs (\cref{fig:exp-image-edits-comparisons}).

\paragraph{Character Insertion.}
The task spec above is closely relevant to the character insertion task studied in Magic Insert~\citep{ruiz2024magic}, where the goal is to insert a character from an image into a background image. We use 10 background images paired with 8 character inputs released on their official demo page to construct an evaluation dataset of 80 scenes. 
Since their model uses a different backbone~\citep{podell2023sdxl}, for fair comparisons, we also include three baselines using the same FLUX backbone as ours: Add-it~\citep{tewel2024add}, SDEdit~\citep{meng2021sdedit}, and FLUX-Fill. 

Results are in \cref{tab:editing} (bottom half). Add-it receives text instructions for object insertion, but tends not to follow the specified character location in the prompts; SDEdit produces harmonized outputs but does not closely preserve the background due to the global noising operation; the inpainting method, FLUX-Fill, sometimes ignores the character descriptions in the input text instruction and fails to insert any object. These observations are also reflected qualitative samples in \cref{supp:exp-editing}.

\begin{table}[t]
\centering
\scriptsize
\setlength{\tabcolsep}{0.7pt}
\begin{tabular}{lcccccccccc}
\toprule
\multirow{2}{*}{Methods} & \multicolumn{3}{c}{Controllability} & \multicolumn{4}{c}{Video Quality} &
 \multicolumn{2}{c}{Semantic Alignment}
 \\
\cmidrule(lr){2-4} \cmidrule(lr){5-8}\cmidrule(lr){9-10}
 &IoU (fg) ($\uparrow$) & LPIPS ($\downarrow$) & GPT‑4o ($\uparrow$) & Smooth ($\uparrow$) &
   Aesthetic ($\uparrow$) & Imaging ($\uparrow$)& GPT‑4o ($\uparrow$)  &
   ViCLIP ($\uparrow$)& GPT‑4o ($\uparrow$) \\
\midrule
\multicolumn{9}{l}{\textit{Object-Centric Simulation Input}} \\
\midrule
Traj2V  
&$0.602$\xpm{.193}
& $0.109$\xpm{.033} & 
$0.638$\xpm{.172} & $\underline{0.993}$\xpm{.003} & $0.534$\xpm{.071} & $0.570$\xpm{.154} & 
$0.725$\xpm{.179} & $0.258$\xpm{.041} &  $0.757$\xpm{.141} \\
Depth2V 
&$\mathbf{0.787}$\xpm{.229}
& $\mathbf{0.103}$\xpm{.031} & 
$\underline{0.650}$\xpm{.144} & $0.984$\xpm{.015} & $0.534$\xpm{.069} & $0.581$\xpm{.155} &
$0.750$\xpm{.126} & $\underline{0.261}$\xpm{.041} &  $0.775$\xpm{.130} \\
Image2V 
&$0.321$\xpm{.249}
& $0.111$\xpm{.021} & 
$\mathbf{0.708}$\xpm{.104} & $0.978$\xpm{.025} & $\mathbf{0.553}$\xpm{.057} & $\underline{0.611}$\xpm{.101} & 
$\underline{0.775}$\xpm{.099} & $0.255$\xpm{.036} &  $\underline{0.788}$\xpm{.079} \\
Ours           
&$\underline{0.739}$\xpm{.221}
& $\underline{0.104}$\xpm{.030} &
$\mathbf{0.708}$\xpm{.104} & $\mathbf{0.994}$\xpm{.003} & $\underline{0.549}$\xpm{.068} & $\mathbf{0.625}$\xpm{.117} &
$\mathbf{0.842}$\xpm{.132} & $\mathbf{0.270}$\xpm{.038} &  $\mathbf{0.817}$\xpm{.080} \\
\midrule
\multicolumn{9}{l}{\textit{Full-Scene Simulation Input} (PhysGen3D~\citep{chen2025physgen3d} dataset)} \\\midrule
PhysG3D &--& -- & $0.438$\xpm{.099} & $\mathbf{0.995}$\xpm{.002} & $0.571$\xpm{.114} & $0.673$\xpm{.046} & $0.513$\xpm{.117} & $0.224$\xpm{.026} & $0.588$\xpm{.117} \\
Inversion&--& $0.237$\xpm{.084} & $0.263$\xpm{.122} & $\underline{0.993}$\xpm{.002} & $0.468$\xpm{.097} & $0.410$\xpm{.144} & $0.250$\xpm{.206} & $0.202$\xpm{.058} & $0.400$\xpm{.206} \\
Depth2V &--& $\underline{0.164}$\xpm{.065} & $\underline{0.550}$\xpm{.173} & $\underline{0.993}$\xpm{.003} & $\underline{0.579}$\xpm{.067} & $\underline{0.680}$\xpm{.081} & $\underline{0.662}$\xpm{.216} & $\mathbf{0.242}$\xpm{.030} & $\underline{0.788}$\xpm{.108} \\
Image2V &--& $0.242$\xpm{.077} & $0.525$\xpm{.192} & $0.986$\xpm{.010} & $\mathbf{0.594}$\xpm{.074} & $0.662$\xpm{.052} & $0.638$\xpm{.245} & $0.236$\xpm{.028} & $\underline{0.788}$\xpm{.105} \\
Ours &--& $\mathbf{0.136}$\xpm{.047} & $\mathbf{0.587}$\xpm{.220} & $\underline{0.993}$\xpm{.003} & $0.575$\xpm{.071} & $\mathbf{0.688}$\xpm{.045} & $\mathbf{0.763}$\xpm{.132} & $\underline{0.239}$\xpm{.038} & $\mathbf{0.825}$\xpm{.141} \\
\bottomrule
\end{tabular}
\vspace{-1.5em}
\caption{\textbf{Physics-Simulator-Instructed Video Generation Evaluation.}}
\vspace{-2.7em}
\label{tab:video}
\end{table}
\begin{figure}
    \centering
    \includegraphics[width=\linewidth]{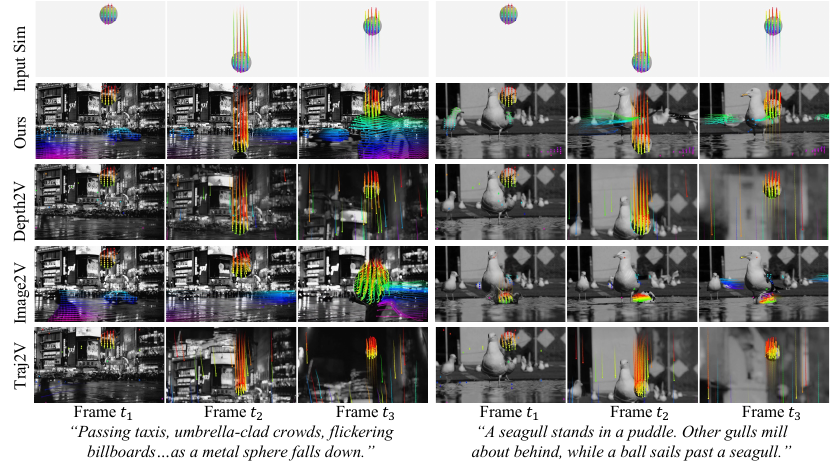}
    \vspace{-2.6em}
    \caption{\textbf{Comparisons on Physics-Simulator-Instructed Video Generation.} Predictions are processed in grayscale and overlaid with estimated tracking~\citep{xiao2024spatialtracker} for visualization.}
    \label{fig:exp-video-tracking}
    \vspace{-1.8em}
\end{figure}

\subsection{Physical-Simulator-Instructed Video Generation}
\label{sect:exp-video}
\paragraph{Task.}
The setup from \cref{sect:exp-editing} can be directly extended to dynamic scenes for video generation, with object poses and dynamics (e.g., a ball bouncing, \cref{fig:exp-video} top row) specified by a physical simulator, an input image, and a textual scene description. Below, we consider two sets of experts: flow-based~\citep{wan2025} and autoregressive-based~\citep{zhang2025framepack}.

\paragraph{Flow-Based Experts.}
We evaluate our method and several monolithic models, including image-to-video (Image2V), depth-to-video (Depth2V), and trajectory-to-video (Traj2V), all using the same Wan2.1 (14B) model as backbone, on a dataset consisting of 3 simulation inputs paired with 4 input images each, giving a total of 12 scenes with examples in \cref{fig:exp-video}. 
To obtain the initial frame for all methods, we edit the input image with our method described in \cref{sect:exp-editing} to insert the simulated object. 
Depth maps required as method inputs are rendered from the physical simulator, and input trajectory maps are computed using object centroids. Metrics include the ones from \cref{sect:exp-editing} that are relevant to this task, with additional ones adopted from a video benchmark VBench~\citep{huang2024vbench}: 
motion smoothness scores introduced in VBench, MUSIQ~\citep{ke2021musiq} for imaging quality, and ViCLIP~\citep{wang2023internvid} for text prompts alignment. We further compute foreground motion trajectory accuracy with IoU on SAM2~\citep{ravi2024sam2} detected from output videos.

Results are in \cref{tab:video} (top half). Depth-to-video and trajectory-to-video alone fail to capture rich non-foreground-object motions, such as cars moving and birds running in \cref{fig:exp-video-tracking}. In particular, they tend to compensate for the object's downward motion with upward camera motions, as visualized by the tracking trajectories in \cref{fig:exp-video-tracking}. Image-to-video model does not follow the input motion.

\begin{figure}[t]
    \centering
    \includegraphics[width=\linewidth]{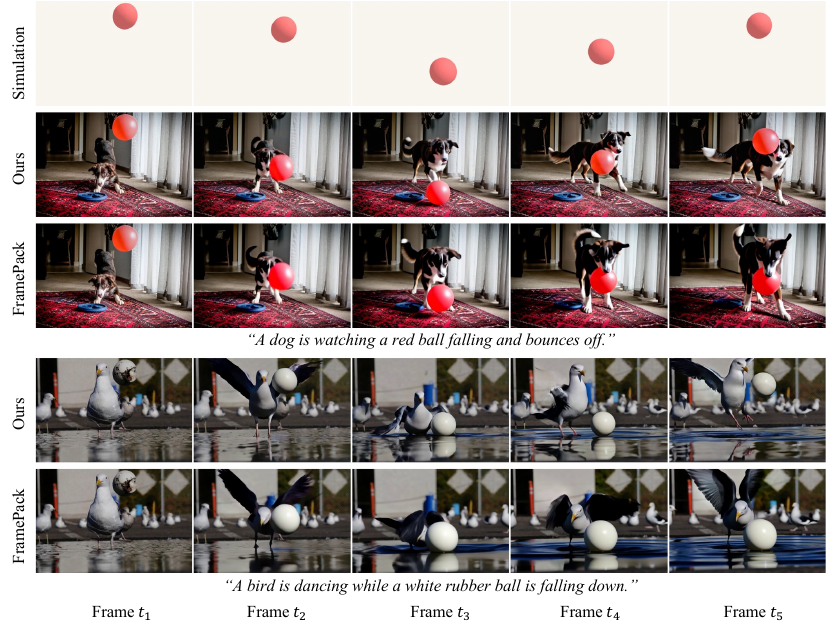}
    \vspace{-2.6em}
    \caption{\textbf{Physical-Simulation-Instructed Video Generation} compared with the baseline with the same backbone. Our method better adheres to the input object motion trajectory.}
    \vspace{-.8em}
    \label{fig:exp-video-ar}
\end{figure}

\paragraph{Autoregressive Experts.}
We implement the autoregressive construction for \cref{eqn:target-generative-annealed} using two generative experts: a next-frame-section video prediction model, FramePack~\citep{zhang2025framepack}, and a Gaussian distribution $p^\text{sim}(x)$ centered on physics simulator renderings. 
Denote a video sequence of the physics simulator's RGB rendering overlayed with the input initial frame as $c^\text{sim}$, then $p^\text{sim}(x) :\propto \exp{(-w\|x - c^\text{sim}\|_2^2)}$ with constant $w\in\mathbb{R}$. 
In this case, $x_{1:t}$ represents the first $T+1-t$ frame sections of a video $x$. We approximate sampling from this distribution with its MAP solution, which amounts to gradient updates w.r.t. loss $\|x - c^\text{sim}\|_2^2$.
Qualitative results are in \cref{fig:exp-video-ar}, suggesting that composing per-expert constraints enforces desired object motion on top of output photorealism. 

\paragraph{Full-Scene Simulation.}
In the above setting, simulation inputs only describe foreground objects. We further evaluate cases where full-scene simulation is available. We use the simulation results from 9 released scenes from PhysGen3D~\citep{chen2025physgen3d}, a method that reconstructs and animates images in physical simulators, as evaluation inputs. 
To better preserve scene content, we use the flow-inverted noise vector for particle initialization as opposed to random noise, and compare with this inversion baseline. Results are included in \cref{tab:video} with qualitative samples in \cref{fig:exp-physgen3d}.

\begin{table}[t]
\centering
\scriptsize
\begin{tabular}{lccc}
\toprule
Method & mIoU ($\uparrow$) & VQAScore ($\uparrow$) & VQAScore-R ($\uparrow$) 
\\
\midrule
3DIS-FLUX
 & $0.673$\xpm{0.182}
 & $0.800$\xpm{0.190}
 & $0.808$\xpm{0.161}
 \\
$L = 1$ (\cite{du2023reduce})
 & $0.675$\xpm{0.200}
 & $0.567$\xpm{0.272}
 & $0.567$\xpm{0.189}
 \\
$L = 4$
 & $0.695$\xpm{0.172}
 & $0.802$\xpm{0.190}
 & $0.795$\xpm{0.155}
 \\
$L = 8$
 & $0.715$\xpm{0.177}
 & $0.879$\xpm{0.115}
 & $0.843$\xpm{0.135}
 \\
$L = 16$
 & \underline{$0.723$}\xpm{0.176}
 & \underline{$0.897$}\xpm{0.106}
 & \underline{$0.870$}\xpm{0.128}
 \\
$L = 32$
 & $\mathbf{0.728}$\xpm{0.170}
 & $\mathbf{0.904}$\xpm{0.092}
 & $\mathbf{0.881}$\xpm{0.115}
 \\
\bottomrule
\end{tabular}
\vspace{-1em}
\caption{\textbf{Text-to-Image Generation with Layout Controls} on MIG-Bench~\citep{zhou2024migc} dataset, comparing our method with varying compute budgets and an application-specific baseline. 
}
\vspace{-2.4em}
\label{tab:layout}
\end{table}

\subsection{Text-to-Image Generation with Layout Control}
\label{sect:exp-image}

\paragraph{Task and Evaluation.}
This task aims to generate an image given an input global text prompt and a set of object bounding boxes with paired object text descriptions.
We evaluate on 50 scenes randomly sampled from the full MIG-Bench~\citep{zhou2024migc} dataset of 800 scenes. 
Metrics include bounding box mIoU, detected using GroundingDINO~\citep{liu2024grounding}, and VQAScore~\citep{lin2024evaluating} measuring output alignment with global and regional prompts.

\paragraph{Results.}
\cref{tab:layout,fig:supp-mig-bench} contains results on ablating the number of particles $L$ and also comparisons with 3DIS-FLUX~\citep{zhou20253dis}, a state-of-the-art method designed for text-to-image generation with the same backbone model as ours. In the case of $L=1$, the method reduces to \cite{du2023reduce}. 
The performance of our method improves with higher computation budgets, and alleviates the need for application-specific designs, such as cross-attention layer intervention as used in \citet{zhou20253dis}. The ablation on SMC resampling is deferred to \cref{app:abl-smc}.

\section{Conclusion}
We have introduced a probabilistic framework where knowledge from heterogeneous sources is composed in the form of product distributions, and proposed an efficient sampling algorithm interleaving annealed MCMC sampling and SMC resampling to integrate guidance from multiple generative and discriminative models. Our training-free approach can effectively integrate visual generative priors, VLM guidance, and physics-based constraints. Empirical evaluation suggests that this method generates images and videos with improved controllability and fidelity compared to prior works, providing a practical recipe for controllable visual synthesis tasks.

\bibliography{main}
\bibliographystyle{iclr2026_conference}

\newpage
\appendix
\section{Overview}
The appendix contains discussions (\cref{app:discussions}), extended descriptions of the framework (\cref{app:framework-details}), and additional experimental results (\cref{app:exp-results}) and details (\cref{app:exp-details}).

\section{Discussions}
\label{app:discussions}

\paragraph{Compute Requirement.} All image experiments are run using 1 NVIDIA H200 GPUs and all video experiments with 2 NVIDIA H200 GPUs. 
Generating one sample in \cref{sect:exp-editing} takes approximately 4min. It takes 5min for flow-based and 30min for autoregressive-based models in \cref{sect:exp-video}.
Wall-clock time for the task in \cref{sect:exp-image} with different numbers of particles is included in \cref{tab:supp-abl-smc}.

\paragraph{Limitations.} 
The proposed method requires extra computation due to intermediate MCMC steps and parallel sampling compared to vanilla feedforward inference of the image or video generative model backbones. 
Furthermore, the sampling efficacy relies on the assumption that expert predictions are reasonably compatible, and strong incompatibilities might require other sampling techniques or training. This work proposes a framework and provides empirical validation with small number of experts and we leave large-scale sampling as future work. 

\paragraph{Societal Impact.}
We believe outputs of the proposed framework does not directly possess negative societal impact. The underlying technology enable more content creation tools and benefits understanding model behaviors. 
However, highly controllable generation could be exploited to create misinformation. While our work does not inherently use sensitive data, we are aware of such risks and closely follow ethical guidelines in the community to help mitigate these risks.

\section{Framework Details}
\label{app:framework-details}

\subsection{Flow Models}
\label{app:framework-details-flow}

We use bar notations for pre-trained models in the following discussions.

Consider the $i$-th generative expert, which is a flow model that defines a time-dependent continuous probability path $\bar{p}^{(i)}(x; \bar{t}), x\in\mathbb{R}^d, \bar{t}\in\left[0, 1\right]$, with boundary conditions $\bar{p}^{(i)}(x; 0) = \mathcal{N}(0, I)$ and $\bar{p}^{(i)}(x; 1) \approx p_\text{data}^{(i)}(x)$. 
Each flow model is associated with a scheduler $\bar{\alpha}, \bar{\sigma}: \left[0, 1\right] \rightarrow \mathbb{R}$ that defines a conditional probability path $\bar{p}^{(i)}(x\mid y; \bar{t}) = \mathcal{N}(x\mid \bar{\alpha}(\bar{t}) y, \bar{\sigma}(\bar{t}) I), y\sim p_\text{data}^{(i)}(y)$, and is trained to predict a velocity 
field $\bar{v}^{(i)}(x; \bar{t}, \theta_i)$ that generates the marginalized path $\bar{p}^{(i)}(x;\bar{t})$, where $\theta_i$ denotes model parameters. 
While the training-time scheduler $(\bar{\alpha}, \bar{\sigma})$ can differ across experts, they can be aligned during inference time via a scale-time transformation~\citep{lipman2024flow}. 

Given a time remapping function $\xi: \{T, T-1, \cdots, 1\} \rightarrow \left[0, 1\right]$, one can construct the per-expert product component from \cref{eqn:target-generative-annealed} via discretizing the path $\bar{p}^{(i)}(x; \bar{t})$ into 
\begin{equation}
    p^{(i)}_t(x):= \bar{p}^{(i)}(x; \xi(t)).
    \label{eqn:instantiated-generative-annealed}
\end{equation}
We require that $\xi$ is monotonic and $\xi(T) = 0, \xi(1) = 1$. By construction, $p^{(i)}_T(x) = \mathcal{N}(0, I)$ is easy to sample from, and $p^{(i)}_1(x) = p_\text{data}^{(i)}(x)$ encapsulates the prior data distribution of the expert model, fulling our motivation in \cref{sect:generative_expert}. 
For all experiments, we define $\xi$ to be the Euler discretization~\citep{esser2024scaling,euler1792institutiones}, with additional alignment to the DDPM scheduler~\citep{ho2020denoising} for experiments in \cref{sect:exp-image}. 

Line 5 from \cref{alg:implementation} is computed as follows:
\begin{align*}
    v_t^{(i)}(x) &= \bar{v}^{(i)}(x; \xi(t), \theta_i), \\
    \alpha_t &= \bar{\alpha}(\xi(t)), \\
    \sigma_t &= \bar{\sigma}(\xi(t)), \\
    s_t^{(i)}(x) &= \frac{-\alpha_t v_t^{(i)}(x) + \dot{\alpha}_t x}{\dot{\sigma_t}\sigma_t\alpha_t - \dot{\alpha}_t \sigma^2_t},\\
    v_t(x) &=\sum_i \lambda^{(i)}(x) v^{(i)}(x), \\
    s_t(x) &= \sum_i \lambda^{(i)}(x) s^{(i)}(x), \\
    \mathcal{K}_{t\leftarrow t+1}(x^\prime\mid x) &= \delta(x^\prime - (x +v_t(x) (\xi(t-1) - \xi(t))  )), \\
    \mathcal{K}_t(x^\prime\mid x) &= \mathcal{N}(x^\prime; x + \frac{\kappa^2}{2}s_t(x), \kappa^2 I),
\end{align*}
where $\lambda^{(i)}(x)$ is defined in \cref{app:exp-details-overall}, with all entries of $\sum_i \lambda^{(i)}(x)$ to be $1$. 

The end-point prediction is available as
\begin{equation}
    \hat{x} = \frac{\sigma_t}{\dot{\alpha}_t \sigma_t - \dot{\sigma}_t \alpha_t} v_t(x) - \frac{\dot{\sigma}_t}{\dot{\alpha}_t \sigma_t - \dot{\sigma}_t \alpha_t} x.
    \label{eqn:flow-endpoint}
\end{equation}

\subsection{Framework Instantiation with Autoregressive Experts}
\label{app:framework-details-autoregressive}

\subsection{Approximation for Conditional Sampling}
\label{app:conditional-sampling}
We explain \cref{eqn:conditional-update} below.
Let $x_i$ be any expert. We approximate the distribution $p(x_i \mid x_{\text{pa}(i)})$ as
\begin{equation}    
p(x_i \mid x_{\text{pa}(i)}) \propto p(x_i) p(x_{\text{pa}(i)} \mid x_i) = p(x_i)  \prod_{i^\prime\in\text{pa}(i)} p(x_{i^\prime} \mid x_i),
\label{eqn:appendix_compose_conditional}
\end{equation}
 where parent regions $x_{i^\prime}$ are assumed to be conditionally independent given $x_i$. To model each conditional distribution $p(x_{i^\prime}\mid x_i)$, we define a Gaussian distribution of the deviation of the flow vector predicted at $x_{i^\prime}$ and $x_i$, where $p(x_{i^\prime}\mid x_i) \propto e^{-w \| v(x_{i^\prime}) - v(x_i) \|^2}$ with a constant $w\in\mathbb{R}$. Under this parametric distribution, when $x_{i^\prime}$ and $x_i$ have consistent flow predictions, the likelihood is high (which is reasonable as this indicates both $x_{i^\prime}$ and $x_i$  are mutually compatible). We can convert the probability expression in \cref{eqn:appendix_compose_conditional} to a corresponding score function, resulting in \cref{eqn:conditional-update}. 

In this work, $\text{pa}(i)$ is defined as the generative expert defining a natural data distribution for global $x\in \mathbb{R}^d$, e.g., a text-conditioned generative model for the full image/video $x$. 
Each generative expert is additionally conditioned on a context signal $c_i\in\mathbb{R}^{d_i^\text{context}}$, e.g., texts or depth maps. 
Note that the framework allows for different types of conditions across experts, enabling flexible control handles that are typically application-dependent.

\section{Extended Experimental Results}
\label{app:exp-results}

\subsection{Graphics-Engine Instructed Image Editing}
\label{supp:exp-editing}
\begin{figure}[t]
    \centering
    \includegraphics[width=\linewidth]{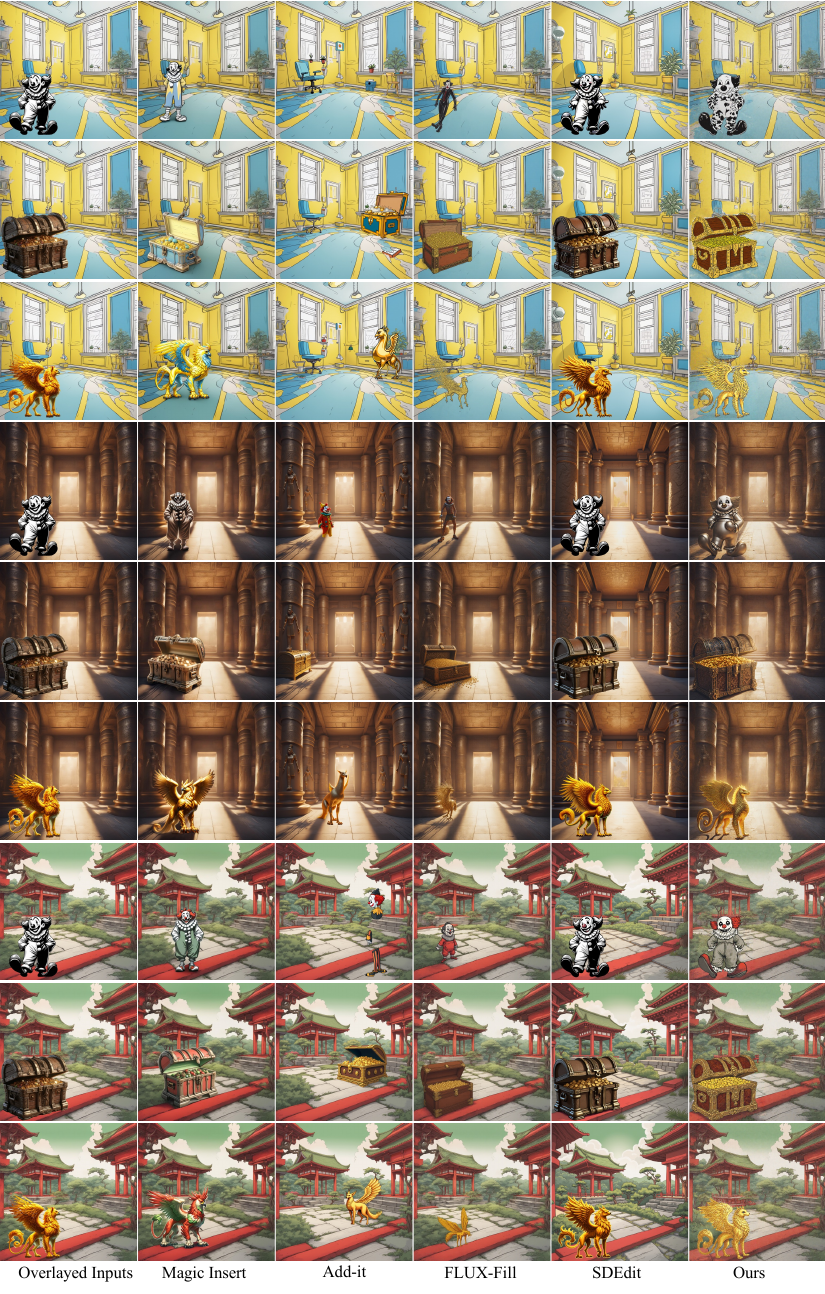}
    \vspace{-2.8em}
    \caption{\textbf{Comparisons on Character Insertion Task.}}
    \vspace{-1em}
    \label{fig:supp-magic-insert}
\end{figure}

\cref{fig:supp-magic-insert} contains qualitative samples on the Magic Insert dataset for experiments in \cref{sect:exp-editing}.

\subsection{Physical-Simulator-Instructed Video Generation}

\begin{figure}
    \centering
    \includegraphics[width=\linewidth]{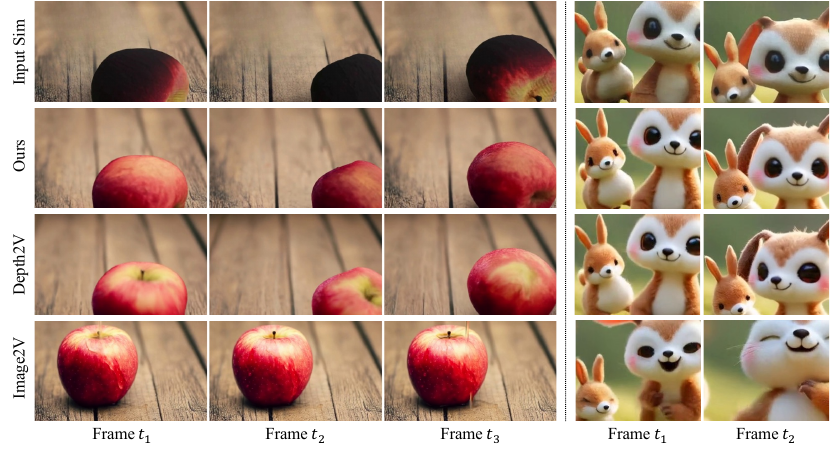}
    \caption{\textbf{Baseline Comparisons on PhysGen3D Data.} Our method improves the visual quality compared to the input simulation and adheres more closely to the input object motions compared to baselines.}
    \label{fig:exp-physgen3d}
\end{figure}

Qualitative samples are included in \cref{fig:exp-physgen3d}.

\subsection{Text-to-Image Generation with Regional Composition}

\label{supp:exp-image}
\begin{figure}[t]
    \centering
    \includegraphics[width=\linewidth]{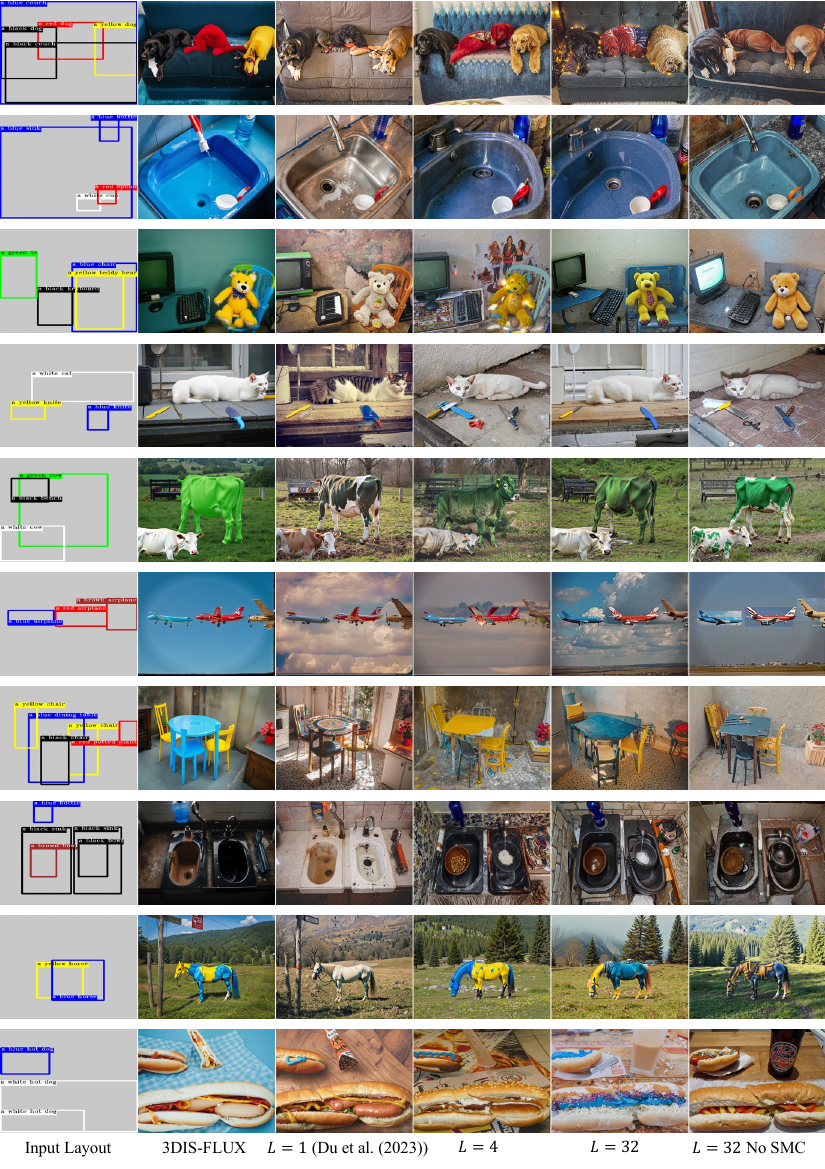}
    \caption{\textbf{Comparisons on Layout-Conditioned Image Generation Task.}}
    \label{fig:supp-mig-bench}
\end{figure}

\cref{fig:supp-mig-bench} contains qualitative samples on the MIG-Bench dataset for experiments in \cref{sect:exp-image}. 

\subsection{Ablation on SMC}

\label{app:abl-smc}

\begin{table}[t]
\centering
\scriptsize
\begin{tabular}{lcccc}
\toprule
Method & mIoU ($\uparrow$) & VQAScore ($\uparrow$) & VQAScore-R ($\uparrow$) & Wall-Clock Time (min)
\\
\midrule
$L = 1$
 & $0.675$\xpm{0.200}
 & $0.567$\xpm{0.272}
 & $0.567$\xpm{0.189}
 & $\sim$1
 \\
$L = 1$ (No SMC)
  & $0.675$\xpm{0.200} ($+0.000$)
 & $0.567$\xpm{0.272} ($+0.000$)
 & $0.567$\xpm{0.189} ($+0.000$)
  & $\sim$1
 \\
$L = 4$
 & $0.695$\xpm{0.172}
 & $0.802$\xpm{0.190}
 & $0.795$\xpm{0.155}
  & $\sim$4
 \\
$L = 4$ (No SMC)
  & $0.715$\xpm{0.170} (\plus{+0.020})
 & $0.795$\xpm{0.170} (\minus{-0.007})
 & $0.767$\xpm{0.159} (\minus{-0.028})
   & $\sim$4
 \\
$L = 8$
 & $0.715$\xpm{0.177}
 & $0.879$\xpm{0.115}
 & $0.843$\xpm{0.135}
   & $\sim$6
 \\
$L = 8$ (No SMC)
   & $0.735$\xpm{0.163} (\plus{+0.020})
 & $0.807$\xpm{0.182} (\minus{-0.072})
 & $0.795$\xpm{0.153} (\minus{-0.048})
    & $\sim$6
 \\
$L = 16$
 & \underline{$0.723$}\xpm{0.176}
 & \underline{$0.897$}\xpm{0.106}
 & \underline{$0.870$}\xpm{0.128}
    & $\sim$12
 \\
$L = 16$ (No SMC)
   & $0.698$\xpm{0.177} (\minus{-0.025})
 & $0.814$\xpm{0.186} (\minus{-0.083})
 & $0.806$\xpm{0.162} (\minus{-0.064})
     & $\sim$12
 \\
$L = 32$
 & $\mathbf{0.728}$\xpm{0.170}
 & $\mathbf{0.904}$\xpm{0.092}
 & $\mathbf{0.881}$\xpm{0.115}
     & $\sim$25
 \\
$L = 32$ (No SMC)
   &$0.691$\xpm{0.183} (\minus{-0.037})
 & $0.854$\xpm{0.139} (\minus{-0.050})
 & $0.838$\xpm{0.136} (\minus{-0.043})
      & $\sim$25
 \\
\bottomrule
\end{tabular}
\caption{\textbf{SMC resampling ablations} on the task from \cref{sect:exp-image}.
Values in \textcolor{Green}{green} (\textcolor{Red}{red}) indicate ``No SMC'' is better (worse). 
Results suggest that SMC benefits sampling efficiency. Resampling has relatively marginal computation. 
}
\label{tab:supp-abl-smc}
\end{table}

We conduct ablation for SMC with results shown in \cref{tab:supp-abl-smc}.

\subsection{Text-to-Image Generation}

\begin{figure}
    \centering
    \includegraphics[width=\linewidth]{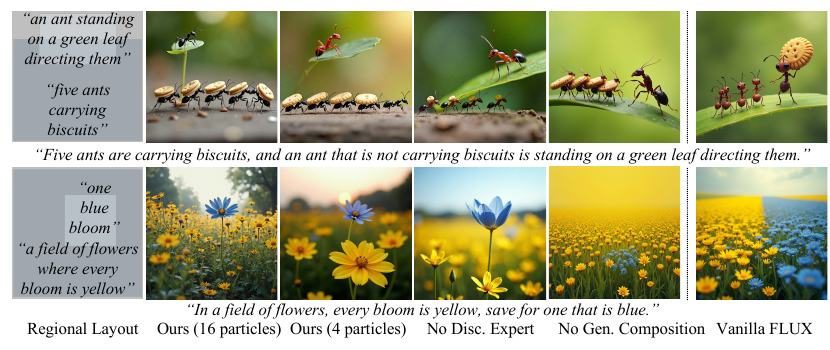}
    \vspace{-1.8em} 
    \caption{\textbf{Text-to-Image Generation}.}
    \label{fig:exp-image}
\end{figure}

We further demonstrate the efficacy of distribution composition on the task of text-to-image generation, where we first query an LLM to parse an input text prompt into spatial bounding boxes with regional prompts (\cref{fig:exp-image} left), aiming to compose the distributions from text-to-image model (FLUX.1 [dev]) conditioned on regional text prompts, one for each region, as generative experts, and from VQAScore~\citep{lin2024evaluating} as a discriminative expert, which gives a scalar value predicting the probability that its input image is aligned with the input text prompt. We compare our methods using $L=16, 4$ particles and $1$ particle (which amounts to not using the discriminative expert), a variant using no generative composition (equivalent to best-of-N sampling for vanilla FLUX) while matching the compute budget for 16 particles, and with the vanilla FLUX backbone. In the examples shown in \cref{fig:exp-image}, our compositional approach achieves better text alignment compared to the baselines, and the discriminative expert provides crucial knowledge, e.g., counting, to steer the output distribution.

\begin{figure}
    \centering
    \includegraphics[width=\linewidth]{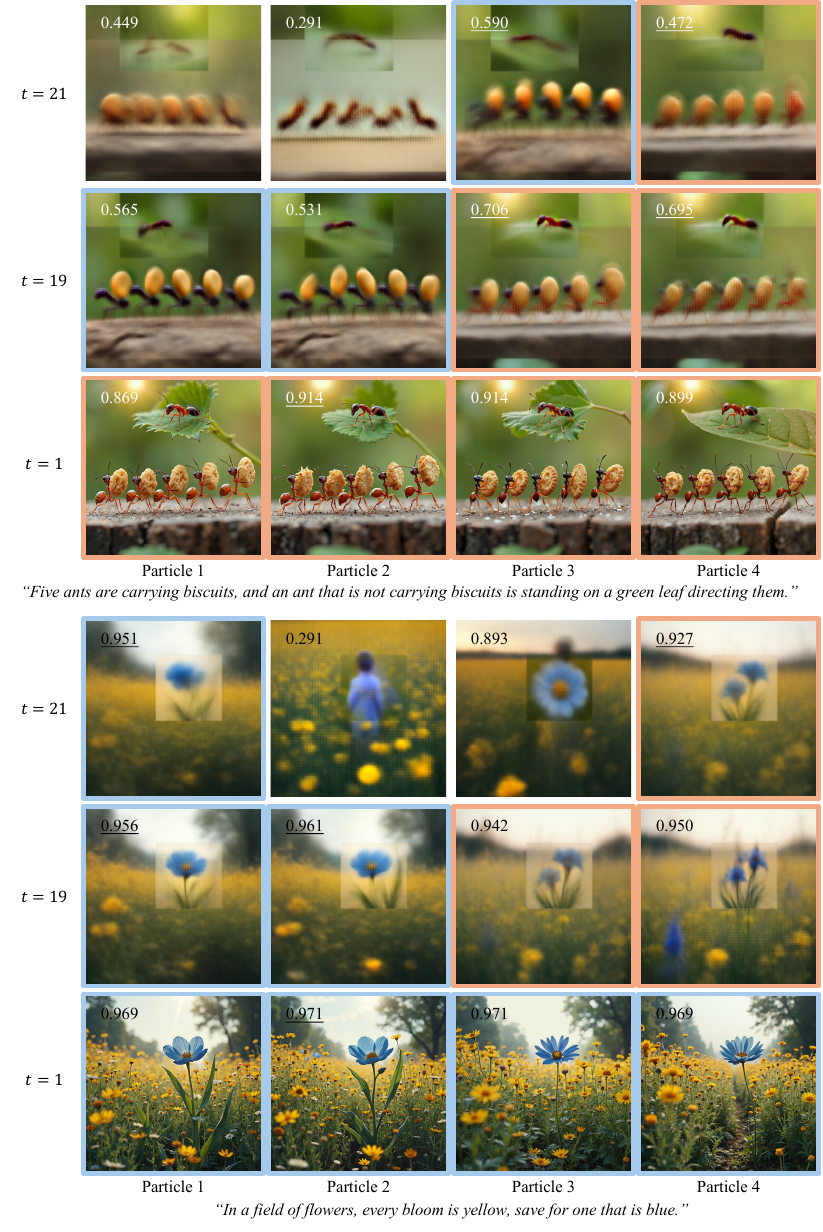}
    \vspace{-1.5em}
    \caption{\textbf{Visualizations of Parallel Sampling with Discriminative Experts.} Discriminative expert (VQAScore) scores are annotated on the top-left corners; score underlining means the sample proceeds to the next annealed distributions (with smaller $t$). Images with the same boundary color share the same initial seed.}
    \vspace{-2em}
    \label{fig:app-image-parallel}
\end{figure}
The parallel sampling process is visualized in \cref{fig:app-image-parallel}. For each scene from \cref{sect:exp-image}, we use an LLM to convert a text prompt into a layout of bounding boxes following the procedure described in \cref{app:exp-details-image}. The input text prompts and resulting layouts and regional prompts are visualized in \cref{fig:exp-image}. 
Then, we follow \cref{alg:implementation} and initialize $L=4$ particles and annealed schedule length $T=28$, which are reweighed at three key timesteps $t=21, 19, 1$ using the discriminative expert, VQAScore~\citep{lin2024evaluating}, which measures image-text alignment with a probability score. At $t=21, 19$, we binarize the scores with the threshold set to be the median score of the current 4 particles, discard the two below the threshold, and duplicate the remaining two high-score particles two times to maintain 4 particles in each round. At $t=1$, we keep the highest-scoring particle as the final output sample. 

\cref{fig:app-image-parallel} shows end-point predictions to visualize the intermediate samples from the annealing process from $t=28$ to $t=1$. VQAScore measurements (higher means more aligned) are overlayed on the top-left corner for each sample. 
In the proposed framework, integrating multiple generative experts allows for composing simpler regional prompts, construction of the annealed distributions allows for iterative refinement of samples, and parallel sampling further integrates the knowledge from the discriminative expert and improves output accuracy, as evidenced in the comparisons from \cref{fig:exp-image}.

\section{Experiment Details}
\label{app:exp-details}

\subsection{Implementation Details}
\label{app:exp-details-overall}

As described in \cref{eqn:target-generative-conditional}, a generative expert can be defined over the entire scene $x\in\mathbb{R}^d$ or a region $x_i\in\mathbb{R}^{d^{(i)}}$ with $d^{(i)}\leq d$, in which case the expert has prior over only part of the scene, e.g., a spatial region in \cref{sect:exp-image}. In these cases, we zero-pad the model prediction from $\mathbb{R}^{d^{(i)}}$ to $\mathbb{R}^d$. 

In other cases, one may restrict the influence of a generative expert to update only a scene region, even when the expert is defined over the full scene $x\in\mathbb{R}^d$, e.g., when the expert only has knowledge over foreground object informed by the graphics engine or physics simulators (\cref{sect:exp-editing,sect:exp-video}). We apply a regional (spatial for images and spatio-temporal for videos) weight $\lambda^{(i)}(x)\in\left[0, 1\right]^d$ on the velocity prediction $v_t^{(i)}(x)$ in these cases, where $\lambda^{(i)}(x)$ is a Gaussian-blurred foreground mask.

\subsection{Automatic Evaluation with VLMs}
\label{app:exp-details-evaluation}
We follow the protocol proposed in PhysGen3D~\citep{chen2025physgen3d} to run automatic evaluation for \cref{sect:exp-image,sect:exp-video} using GPT-4o~\citep{hurst2024gpt4o}. For each scene, all methods' output images, or uniformly sampled 10 frames when outputs are videos, are sent to GPT-4o together with template prompts as specified in PhysGen3D~\citep{chen2025physgen3d}. 
GPT-4o is asked to give scores from $0$ to $1$ for each video on all metrics. The scores for the same metric are averaged across all scenes. 

\subsection{Graphics-Engine-Instructed Image Editing}
\label{app:exp-details-image-edits}
In \cref{sect:exp-editing}, the product distribution is defined as $p(x) \propto p^\text{depth2image}(x\mid c^\text{text}, c^\text{depth}) p^\text{imagefill}(x\mid c^\text{text}, c^\text{image}, c^\text{mask})$. We use $T=28, L=1$, with 1 MCMC sampling step and 2 conditional sampling updates (\cref{eqn:conditional-update}) per iteration. The spatial resolution is $960 \times 1664$ for the graphics engine dataset and $1024 \times 1024$ for Magic Insert dataset. 

For the graphics engine dataset, assets are rendered in Genesis\footnote{\url{https://github.com/Genesis-Embodied-AI/Genesis}}.

For Magic Insert data preprocessing, we use SAM2~\citep{ravi2024sam2} to obtain the segmentation masks of foreground characters, and feed the images of segmented characters, resized to the bottom-left quadrant and overlaid with backgrounds, into GPT o4-mini for image captions to serve as text prompts for evaluated methods.
Magic Insert results are downloaded from the official demo page. Add-it receives background images from the dataset as inputs, together with the GPT-generated captions (of overlaid images) appended with ``\{character\} is at the bottom left quarter'' to indicate the desired locations. SDEdit takes in input overlaid images and adds Gaussian noise with standard deviation $0.5$ before denoising.

\subsection{Physical-Simulator-Instructed Video Generation}
In \cref{sect:exp-video}, for flow models, the product distribution is defined as $p(x) \propto p^\text{depth2video}(x\mid c^\text{text}, c^\text{depth}) p^\text{image2video}(x\mid c^\text{text}, c^\text{image})$. We use $T=30, L=1$ with 1 MCMC sampling step per iteration. We use no conditional sampling updates for this task due to GPU memory constraint. 
Each generated video has $81$ frames and spatial resolution $480 \times 832$. 

For autoregressive models, the product distribution is defined as $p(x) \propto p^\text{sim}(x \mid c^\text{sim})p^\text{image2video}(x\mid c^\text{text}, c^\text{image})$ with $p^\text{sim}(x \mid c^\text{sim})$ defined in \cref{sect:exp-video}. 
We approximate sampling from this distribution with first sampling from $p^\text{image2video}$, then do 8 gradient updates with respect to $\mathcal{L}_2$ loss $\|x - c^\text{simulation}\|^2$.
Results in \cref{fig:exp-video-ar} use $81$ frames per sequence and spatial resolution $512\times 768$. We set $L=1$, with 1 MCMC sampling step per iteration for this experiment. 

\subsection{Text-to-Image Generation with Regional Control}

\label{app:exp-details-image}

For \cref{sect:exp-image}, the product distribution is $p(x) \propto \left(\prod_i p^\text{depth2image}(x_i\mid c^\text{text}, c^\text{depth})\right) q^\text{VLM}(x\mid c^\text{text})$.
Here, $p^\text{depth2image}$ is a regional generative experts (FLUX-Depth), which predict scores for bounding-box-cropped images conditioning on regional text prompts and on regional depth maps cropped from global depth maps predicted using the first stage of 3DIS-FLUX~\citep{zhou20253dis} (layout-to-depth generation); $q^\text{VLM}$ a discriminative expert VQAScore, where for each image sample, it assigns the summed score computed over a full image and its regions cropped with input bounding boxes as the reward.
We use $T=28$ and image resolution $1024 \times 1024$, with $1$ MCMC sampling step and $2$ conditional sampling steps per iteration.

\end{document}